\documentclass[letterpaper, 10 pt, conference]{ieeeconf}  

\IEEEoverridecommandlockouts                              

\usepackage{graphics} 
\usepackage{epsfig} 
\usepackage{times} 
\usepackage{amsmath} 
\usepackage{amssymb}  
\usepackage{amsfonts}  
\usepackage{float}  
\usepackage{stfloats}
\usepackage{graphicx}
\usepackage{booktabs}
\usepackage{makecell}
\usepackage{cite}
\usepackage[hidelinks]{hyperref}  

\title{\LARGE \bf
Road Similarity-Based BEV-Satellite Image Matching for UGV Localization
}

\author{Zhenping Sun, Chuang Yang, Yafeng Bu, Bokai Liu, Jun Zeng and Xiaohui Li \textsuperscript{*}  
	\thanks{Zhenping Sun and Chuang Yang are the co-first authors.} 
	\thanks{* Corresponding author: Xiaohui Li, 152183401@qq.com.} 
	\thanks{The authors are with the College of Intelligence Science and Technology, National University of Defense Technology, Changsha 410073, China.} 
}

\begin{document}

\maketitle
\thispagestyle{empty}
\pagestyle{empty}


\begin{abstract}
	
	To address the challenge of autonomous UGV localization in GNSS-denied off-road environments, this study proposes a matching-based localization method that leverages BEV perception image and satellite map within a road similarity space to achieve high-precision positioning.We first implement a robust LiDAR-inertial odometry system, followed by the fusion of LiDAR and image data to generate a local BEV perception image of the UGV. This approach mitigates the significant viewpoint discrepancy between ground-view images and satellite map. The BEV image and satellite map are then projected into the road similarity space, where normalized cross correlation (NCC) is computed to assess the matching score.Finally, a particle filter is employed to estimate the probability distribution of the vehicle's pose.By comparing with GNSS ground truth, our localization system demonstrated stability without divergence over a long-distance test of 10 km, achieving an average lateral error of only 0.89 meters and an average planar Euclidean error of 3.41 meters. Furthermore, it maintained accurate and stable global localization even under nighttime conditions, further validating its robustness and adaptability.
	
\end{abstract}

\section{INTRODUCTION}

As a core component of autonomous driving systems, localization technology forms the foundation for achieving vehicle autonomous navigation. Current UGV's localization systems typically integrate GNSS modules to provide consistent and reliable global pose estimation. However, in real-world scenarios, GNSS signals may be obstructed by natural or man-made obstacles, leading to temporary failures in the localization system. For UGVs, the current mature localization solution in GNSS-denied environments is based on Simultaneous Localization and Mapping (SLAM) \cite{SLAM} technology. This technique utilizes environmental features obtained through sensors and matches them with the pre-generated map to estimate the current position of the UGV while simultaneously updating the map. In this manner, the UGV is able to perform both map construction and self-localization. However, in the absence of loop closure detection, errors may accumulate over time, leading to a gradual decline in the accuracy of both the map and localization.

In contrast, satellite map is readily accessible and provides a global coverage. Consequently, many approaches leverage satellite map as prior maps, aligning the UGV's perception image with satellite map to determine the position. However, matching the UGV's ground-view images with satellite map for localization presents several challenges, such as the significant viewpoint difference and the lack of environmental features that are typically used for image matching.

\begin{figure}[t]  
	\centering
	\includegraphics[width=\columnwidth]{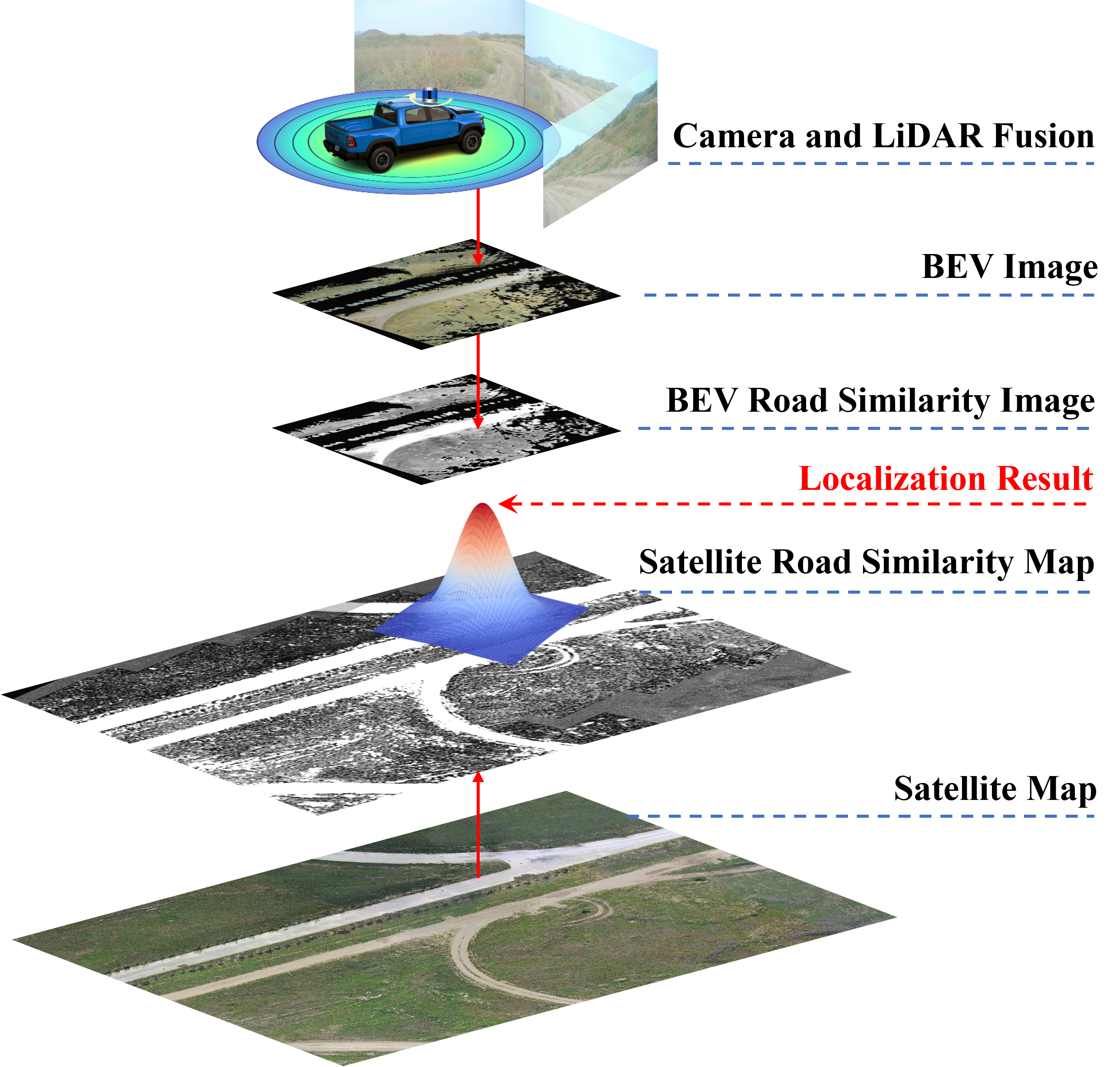} 
	\caption{A diagram of our localization system. The localization is achieved by projecting the BEV perception image, generated through multi-sensor spatiotemporal fusion, and the satellite map into the road similarity space for matching.}
	\label{fig:matching_approach}
\end{figure}

To address these challenges, this paper proposes a new matching approach, as illustrated in Fig.~\ref{fig:matching_approach}, wherein the BEV image and satellite map are transformed into the road similarity space for localization. The proposed localization system consists of four key components:
\begin{enumerate}
	\item First is a high-precision LiDAR and inertial odometry module, which ensures accurate local motion estimation in complex environments. 
	\item Second is the generation of the local BEV representation of the UGV's surrounding environment.
	\item  Third is the road similarity analysis module, which generates both the global satellite road similarity map and the BEV road similarity image.
	\item Fourth is the matching-based localization module, which comprises both image matching and path matching. In the image matching process, a particle filter is employed to estimate the optimal vehicle position, while in the path matching process, the historical motion trajectory is aligned with the global planned path to correct pose errors.
\end{enumerate}

\section{RELATED WORK}

The primary approach to addressing localization issues in GNSS-denied environments is map-based localization using prior maps. The form of the prior map may vary, such as point cloud maps used in LiDAR-based SLAM, aerial and satellite maps, Digital Elevation Models (DEM) \cite{DEM}, and OpenStreetMap (OSM) \cite{OpenStreetMap} data.

With the continuous advancement of remote sensing technology, the resolution of satellite maps has significantly improved in recent years. Consequently, numerous studies have emerged that focus on utilizing prior satellite images to match with vehicle sensor data for localization purposes. Anirudh Viswanathan et al. \cite{viswanathan2014vision} proposed a method in which the panoramic images of the UGV's surroundings are transformed via inverse perspective into a BEV image, aligning the viewpoint with that of the satellite map. They then evaluated the similarity between the BEV and satellite image patch by calculating the Euclidean distance of the SIFT features from both images. Using a particle filter framework for precise localization, their approach demonstrated that satellite map can be effectively utilized for air-to-ground image matching localization. Calvin Cheung et al. \cite{cheung2018cross} applied a similar approach to generate the BEV image; however, their method differs in that, after extracting Canny edge features from both the BEV and satellite map patch, they used fast normalized cross correlation to identify the optimal match between the two. Nevertheless, when the flat ground assumption is violated, both methods generate BEV images with significant artifacts, leading to a degradation in performance.Based on this, et al. Yehonathan Litman \cite{litman2022gps} generate orthographic view images by accumulating geometric features over consecutive frames, while applying probabilistic occupancy grids and filtering techniques to remove noise and artifacts, ensuring the clarity and accuracy of the images. Coupled with NCC, the system is capable of relocalization in GNSS-denied environments. Experimental results demonstrate that the system can perform real-time localization, but due to the limited experimental mileage, the accuracy of its localization for long-distance driving remains unassessed. Kenneth Niles et al. \cite{niles2022satellite} adopted a similar approach. It rasterizes the point cloud into a top-down 2D orthorectified image by mapping the x and y components of each point to an image bin based on the resolution of the satellite image. Localization is subsequently performed using an image template matching method based on NCC.In a 5.1 km experimental route, an average localization error of 1.21 meters was achieved. However, due to hardware issues such as white balance, the generated BEV images failed to accurately reflect the surrounding environment of the vehicle, resulting in poor matching localization performance.

To address the significant cross-view and modality data discrepancies, recent deep metric learning techniques have provided powerful alternatives for cross-view image matching tasks. Researchers have focused on designing robust networks \cite{hu2018cvm,UBEV} that map cross-view images into the same deep embedding space, maximizing the feature similarity between image pairs that are close in the space, while minimizing the similarity between those that are farther apart. For instance, \cite{kim2017satellite} trained a Siamese network that learns an embedding suitable for matching a vehicle's front view image with its corresponding satellite view.This network is designed to distinguish between matching and non-matching ground-satellite pairs by learning discriminative feature representations. The network employs contrastive loss to encourage positive matches to be close in the learned embedding space, while forcing non-matching views to be distant. However, these studies define the satellite map matching problem as an image retrieval task, where the goal is to retrieve the most similar satellite image from a database to determine the query camera's location. In contrast, the aim of our work is to achieve a continuous, stable, and accurate pose estimation.For example, Mengyin Fu et al.  \cite{fu2020lidar} designed a neural network to extract and compare the spatial discriminative feature maps of satellite image patches and LiDAR points, obtaining corresponding probabilities, and then used a particle filter to derive the probability distribution of the vehicle's pose based on the network output.

Some end-to-end approaches have also been proposed. For instance, Laijian Li et al. \cite{2D3D} introduced a Transformer-based 2D-3D matching network, called D-GLSNet, which can directly match LiDAR point clouds with satellite images. Yujiao Shi et al. \cite{boosting3dof} designed a geometry-guided cross-view transformer that maps front-view images to bird's-eye view images. They estimated the relative rotation between the front-view and satellite map by constructing a neural pose optimizer. After aligning their rotations,they then directly regressed the pose offset between the perception image and satellite map.However, these methods suffer from limited generalization ability, which is a common issue in neural network-based learning approaches, particularly when faced with unseen environments or domain shifts.

\section{METHODOLOGY}

\subsection{LiDAR-Inertial Odometry}
The experimental approach integrates LiDAR-inertial odometry (LIO) as the motion model of the UGV. To accurately estimate the relative motion between consecutive time steps, we adopt the FastLio framework \cite{xu2021fast} as the backbone of our LIO system.

\subsection{Generating Local BEV Perception Images via Multi-Modal Spatiotemporal Fusion}

\begin{figure}[htbp]
	\centering
	\includegraphics[width=\columnwidth]{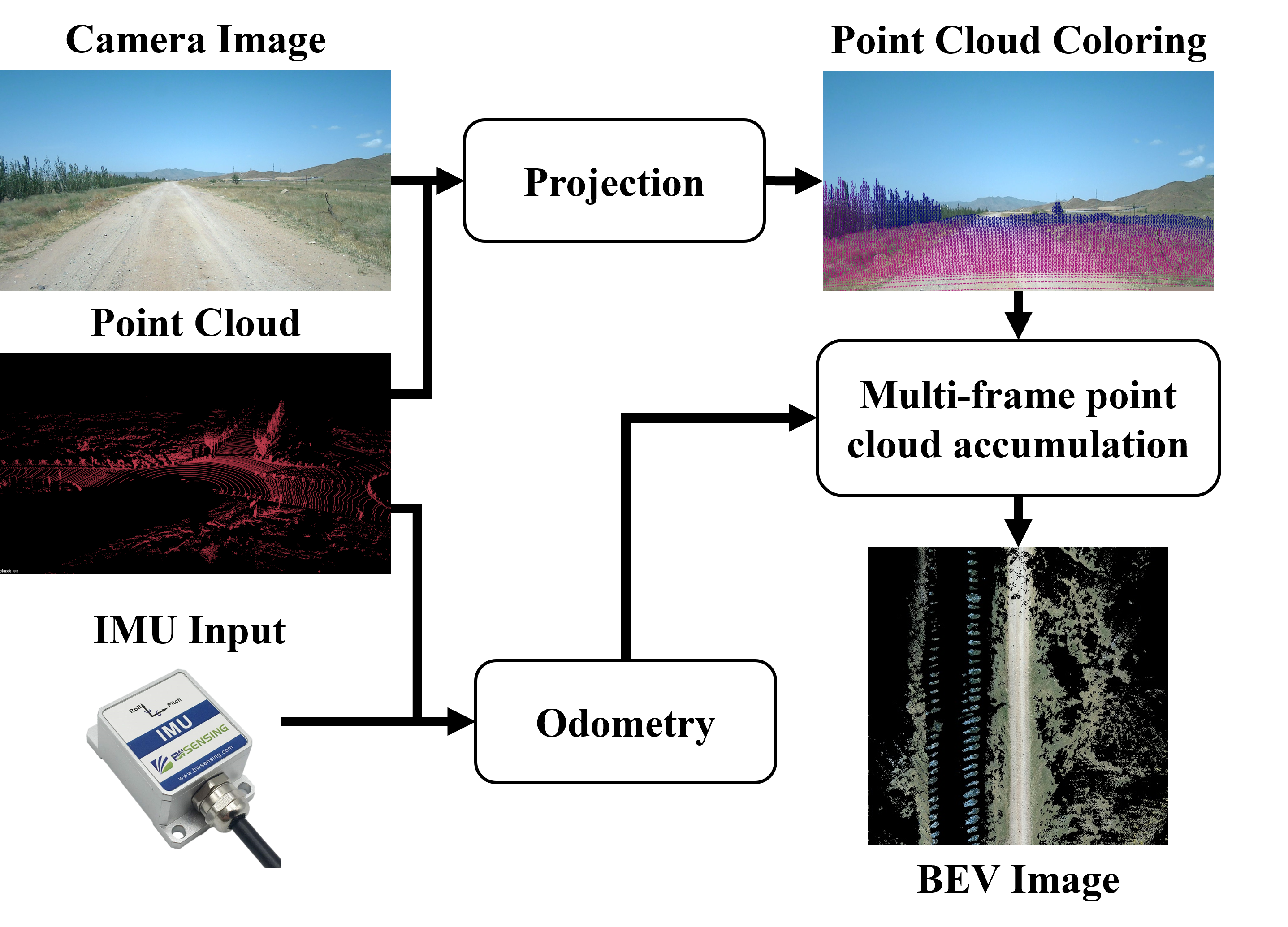}  
	\caption{The bird’s-eye view generation process: For each frame of temporally synchronized point clouds and RGB images, colorize the point clouds using a projection matrix, and accumulate multiple frames of point clouds relying on odometry to complete the bird’s-eye view generation.}
	\label{fig:GetColorMap}
\end{figure}

The process of generating the UGV's local BEV perception image is illustrated in Fig.~\ref{fig:GetColorMap}. After integrating FastLio, we first obtain spatiotemporally aligned data streams through multi-modal fusion, including calibrated LiDAR point clouds, synchronized camera images (visible light or infrared), and tightly coupled LiDAR-inertial odometry pose sequences. Next, using the extrinsic calibration matrix between the LiDAR and the camera, we color the LiDAR point cloud data based on the image data. Subsequently, following the method in \cite{niles2022satellite}, after accumulating multiple frames of LiDAR point cloud data with color information using the pose sequence provided by the odometry data, We rasterize the colored point cloud into a top-down 2D orthorectified image by projecting the x and y coordinates of each point onto the image grid, ensuring alignment with the resolution of the satellite map.The orthorectified image is generated with a resolution of 0.2 meters per pixel, providing sufficient spatial detail to accurately represent the surrounding environment.Finally, we generate a 500$\times$500 pixel Bird's Eye View(BEV) perception image, offering an intuitive representation of the vehicle's surroundings and mitigating the large perspective differences between ground-level observations and satellite map. Furthermore, Under low-light conditions, such as nighttime, the visible-light camera is replaced with an infrared camera. By following the same procedure, a BEV image can still be reliably generated.

\subsection{Road Similarity Analysis}

\begin{figure}[htbp]  
	\centering
	\includegraphics[width=\columnwidth]{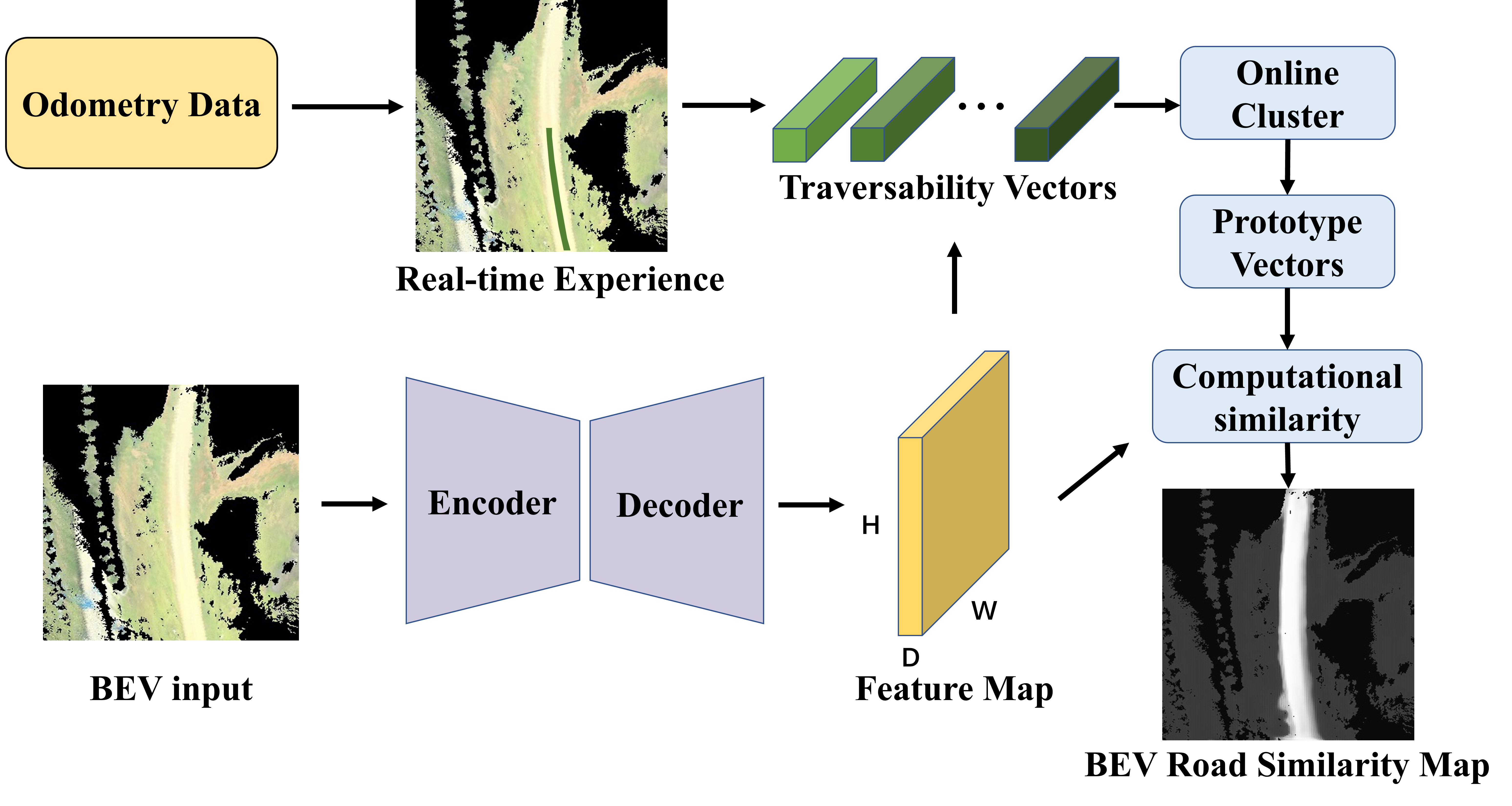}  
	\caption{The road similarity analysis process: First, the feature map is obtained through model inference, and the features of traversable regions are extracted using recorded odometry data. Then, the prototype vectors are analyzed through online clustering, and finally, the traversability map is generated by computing with the feature map.}
	\label{fig:GetSimiMap}
\end{figure}

This study builds upon the work presented in \cite{vstrong}, using the approach of projecting the local BEV perception image and satellite map into a shared feature space, termed the road similarity space, to address the challenges associated with feature matching that are influenced by factors such as geography, season and lighting variations. 

The process of obtaining the BEV road similarity image is illustrated in Fig.~\ref{fig:GetSimiMap}. First, we extract feature maps containing feature vectors for each pixel location from the middle layer of the decoder of an encoder-decoder neural network, which was trained on an image reconstruction task. Next, using the motion trajectory obtained from odometry, we extract the corresponding feature vectors, termed Traversability Vectors, at the pixel locations along the past trajectory in the BEV image. These feature vectors are then subjected to online clustering to obtain prototype vectors that represent road features. Finally, the BEV road similarity image is generated by computing the cosine similarity between all pixel feature vectors and the prototype vectors, which represent road features.

\begin{figure}[htbp]  
	\centering
	\includegraphics[width=0.8\columnwidth]{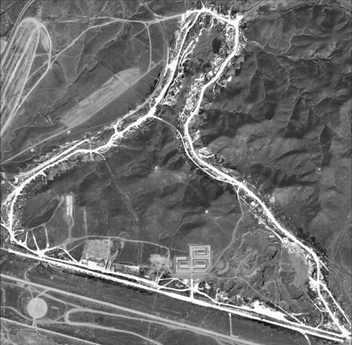} 
	\caption{Global satellite road similarity map.}
	\label{fig:GetSateSimiMap}
\end{figure}

To improve efficiency, the global satellite road similarity map is precomputed offline. The global planned path is first determined either by the global planning module or through manual configuration. Then, a satellite image patch centered around a reference point on the path is extracted. Using the approach outlined in Fig.~\ref{fig:GetSimiMap}, the cosine similarity between the feature vector of each pixel in the patch and the feature vector of the pixels corresponding to the global path is computed, yielding the road similarity result map for the patch. Subsequently, path points are sampled at regular intervals along the global path, and the process is iteratively applied to construct the complete global satellite road similarity map, as illustrated in Fig.~\ref{fig:GetSateSimiMap}.

\subsection{Localization via Matching-Based Estimation}

The matching algorithm in this study consists of two parts: image matching and path matching.

\subsubsection{Image Matching}

\begin{figure}[H]  
	\centering
	\includegraphics[width=0.8\columnwidth]{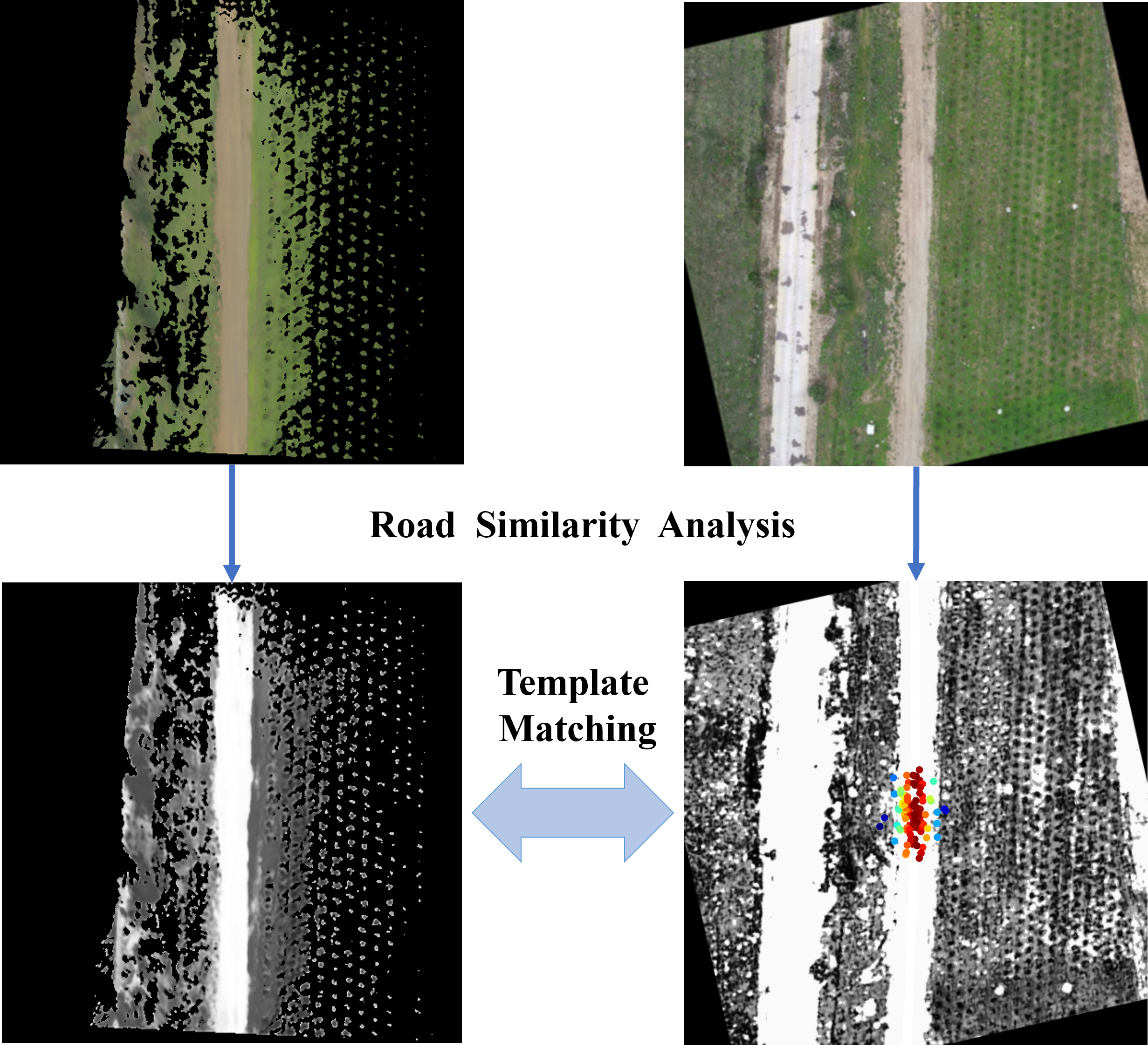}  
	\caption{Image Matching Process: The BEV perception image and the satellite map undergo road similarity analysis. Each particle extracts a satellite map patch from the global road similarity map, using the prior position estimated from odometry as a reference. The NCC between the satellite patch and the BEV road similarity image is then computed to determine the matching score for each particle, where redder colors indicate higher matching scores.}
	\label{fig:TemplateMatching}
\end{figure}

As shown in Fig.~\ref{fig:TemplateMatching}, We employ an image template matching approach to assess the similarity between the BEV road similarity image and a satellite road similarity patch centered at the prior pose. The similarity is quantified using the NCC, defined as:

\begin{equation}
	R(x,y) = \frac{\sum\limits_{i,j} \left( G(i,j) \cdot S(i,j) \right)}
	{\sqrt{\sum\limits_{i,j} G(i,j)^2 \cdot \sum\limits_{i,j} S(i,j)^2}}
	\label{eq:NCC}
\end{equation}

\noindent where $G$ represents the BEV road similarity image, and $S$ denotes the satellite road similarity map patch, which is cropped around the prior position and shares the same dimensions with the BEV image. $R$ represents the NCC value between the two maps. As observed in Eq.~\eqref{eq:NCC}, NCC essentially computes the inner product of the corresponding pixel values in the two images. Due to LiDAR’s susceptibility to occlusions, the BEV perception image often contains unobserved regions (black holes). The NCC computation inherently reduces the impact of these missing areas, improving the robustness of the similarity evaluation.

\begin{figure}[htbp] 
	\centering
	\includegraphics[width=\columnwidth]{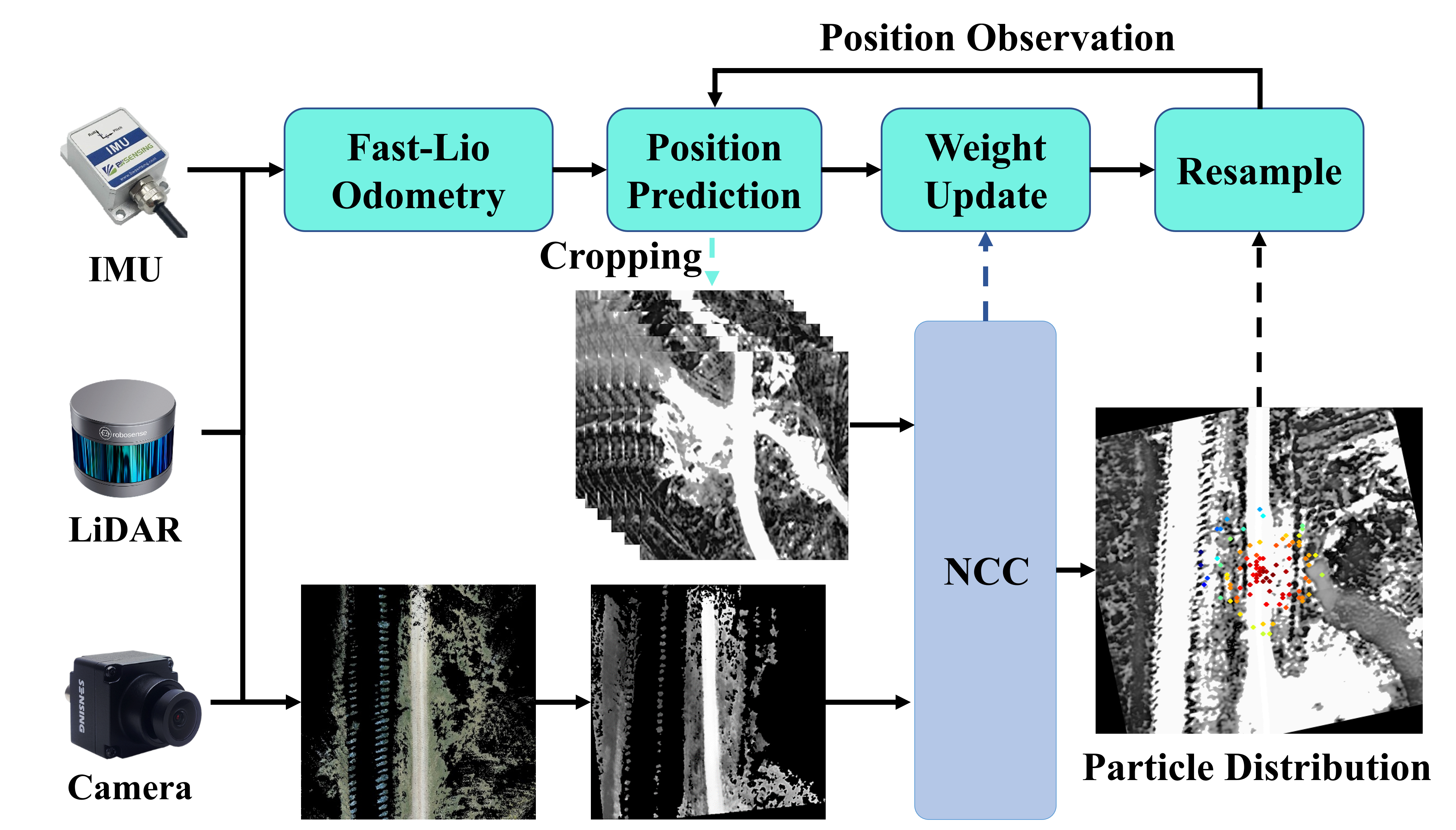} 
	\caption{Illustration of the Particle Filter-Based Localization Method. In our approach, a particle filter is used to estimate the vehicle's pose.}
	\label{fig:PF}
\end{figure}

As illustrated in Fig.~\ref{fig:PF}, we employ a particle filter to estimate the pose of the UGV, representing its probability distribution using a set of weighted particles. Each particle is denoted as $P_{t}^{i}=  (x_{t}^{i}, y_{t}^{i}, \theta_{t}^{i}, w_{t}^{i})$, where \( w_{t}^{i} \) represents the weight of the \( i \)th particle at the time step \( t \), which is derived from the NCC value computed above.
During initialization, if satellite positioning signals are available, they are used to provide an initial pose estimate. Otherwise, we assume that the approximate location of the UGV on the map can be determined through prior knowledge. Given that the proposed localization algorithm is robust to initialization errors, an initial coarse localization can be performed by scanning and matching the BEV road similarity map with the corresponding satellite region to determine a more precise initial position of the UGV.
After position initialization, we randomly sample 100 particles within a 5-meter radius and an orientation range of $ [-1^\circ, 1^\circ] $ around the estimated pose. For each particle, a corresponding patch is extracted from the global satellite road similarity map, and its NCC with the current BEV road similarity image is computed as the particle's weight. Finally, all particle weights are linearly normalized to ensure consistency in the weighting scheme.

During the vehicle motion phase, the relative motion $ (\varDelta Odom_x,\varDelta Odom_y,\varDelta Odom_{\theta})$  between two consecutive frames is predicted using odometry. The prior position of a particle at time $t$ is obtained by adding the predicted relative motion to its pose $L_{t-1}$ at time $t-1$. To account for the uncertainty inherent in odometry-based motion prediction, Gaussian noise is typically introduced during the estimation of the particle’s position at time $t$. The motion update model for each particle at time \( t \) is formulated as follows Eq.~\eqref{eq:MotionEquation}:

\begin{equation}
\left( \begin{array}{c}
	x_{t}^{i}\\
	y_{t}^{i}\\
	\theta _{t}^{i}\\
\end{array} \right) =\left( \begin{array}{c}
	x_{t-1}^{i}\\
	y_{t-1}^{i}\\
	\theta _{t-1}^{i}\\
\end{array} \right) +\left( \begin{array}{c}
	\varDelta Odom_x+N\left( 0,\sigma _{x}^{2} \right)\\
	\varDelta Odom_y+N\left( 0,\sigma _{y}^{2} \right)\\
	\varDelta Odom_{\theta}+N\left( 0,\sigma _{\theta}^{2} \right)\\
\end{array} \right) 
	\label{eq:MotionEquation}
\end{equation}

\noindent where $x_{t}^{i}$ , $y_{t}^{i}$ represents the position in Universal Transverse Mercator (UTM) coordinate, $\theta_{t}^{i}$ represents the heading angle,\( N(0,\sigma) \) represents a noise term with a mean of 0 and a standard deviation of \( \sigma \). After that, we sample a 500$\times$500 pixel satellite image patch centered on the position $x_{t}^{i}$ , $y_{t}^{i}$ with the orientation  $\theta_{t}^{i}$  for each particle.

Then, the NCC between the sampled image and the current BEV road similarity map is computed, and the weight of each particle is obtained using Eq.~\eqref{eq:WeightUpdate}:

\begin{equation}
\tilde{w}_{t}^{i}=w_{t-1}^{i}\cdot NCC\left( G_t,S_{t}^{i} \right) 
	\label{eq:WeightUpdate}
\end{equation}

\noindent where $\tilde{w}_{t}^{i}$ is unnormalized weight at this time step, ${w}_{t-1}^{i}$ is the normalized weight of particle $ {P}_{t-1}^{i} $ at last time step, \( NCC(G,S) \) represents the computation of the NCC between two images, where \( G \) denotes the BEV road similarity map, and \( S_{t}^{i} \) refers to the satellite road similarity map patch, which is cropped around \( P_{t}^{i} \) and has the same dimensions as the BEV image. After normalizing all particle weights, we obtain the final weight distribution at time  \( t \).

If the number of effective particles drops below a predefined threshold, resampling is triggered. This process discards low-weight particles while replicating those with higher weights, improving filter performance. The resampling process is defined as:

\begin{equation}
	N_{eff}=\frac{1}{\sum\limits_{i=1}^N{\left( w_{t}^{i} \right) ^2}}
	\label{eq:EffcetiveParticle}
\end{equation}

Finally, the average of the particles’ pose is calculated to estimate the pose at this time step.

\subsubsection{Global Path Alignment for Localization Refinement}

Given a globally planned path, the estimated motion trajectory over a past time window, obtained from the proposed localization method, can be aligned with the planned path through a combination of rotation and translation.To facilitate alignment, we reformulate the path matching problem as an image registration task. Specifically, the historical motion trajectory and a segment of the globally planned path are separately rendered on black-background images, with both represented as white curves of uniform pixel width. By applying a combination of rotation and translation to the trajectory image, we maximize its overlap with the path image to estimate the optimal transformation parameters. These parameters are then used to correct localization errors at the current time step. To ensure real-time performance and maintain localization consistency, path matching is performed at fixed intervals rather than at every time step.

\section{EXPERIMENTAL VALIDATION}

\subsection{Experiment Setting}

\begin{figure*}[t] 
	\centering
	\includegraphics[width=\textwidth]{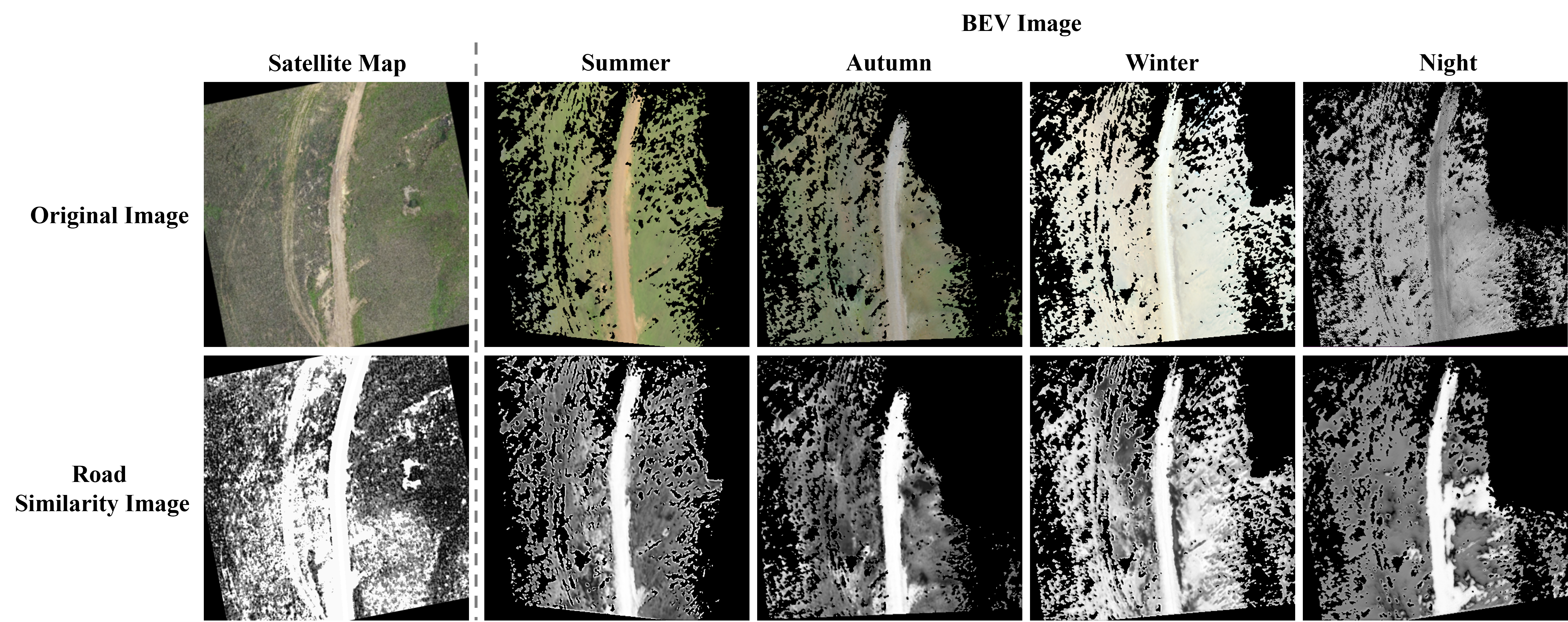}  
	\caption{Similarity analysis between satellite map patches and BEV perception images captured in different seasons and night-time conditions at the same location.}
	\label{fig:SimiContrast}
\end{figure*}

\begin{figure}[htbp]  
	\centering
	\includegraphics[width=\columnwidth]{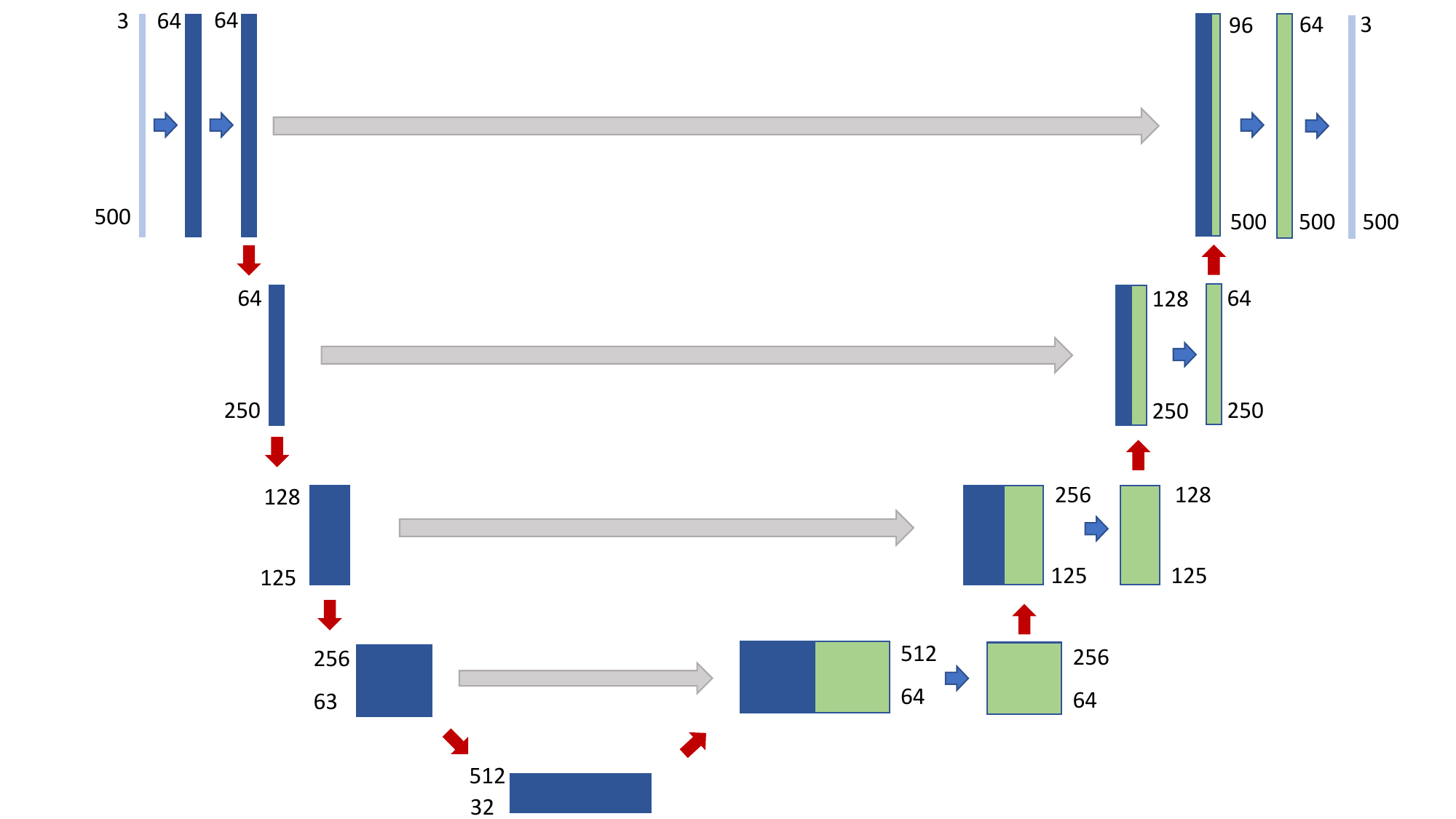}
	\caption{U-Net-based Network.}
	\label{fig:UNet}
\end{figure}

The vehicle-mounted multi-sensor data acquisition system used in this experiment comprises three visible-light cameras, three infrared cameras, a 128-line LiDAR, and an Inertial Measurement Unit (IMU). To ensure the consistency of multi-source data, a rigorous spatio-temporal calibration process was conducted to synchronize all sensors. This process ensures that the data streams of each sensor are temporally aligned and spatially consistent at each sampling moment, providing a reliable reference framework for subsequent multi-modal data fusion and analysis. Image and LiDAR data are recorded at 10 Hz, while the IMU operates at 200 Hz. The generated BEV images are updated at 10 Hz. Additionally, the vehicle is equipped with a dual-antenna GNSS/Inertial Navigation System (INS) to provide ground-truth localization. The reference satellite maps used in this experiment were obtained from the latest 20-level satellite imagery available in BigMap software \cite{bigmap}. To ensure consistency with the BEV image resolution, their pixel resolution was adjusted to 0.2 meters per pixel using bilinear interpolation. Given that the BEV image has a size of 500×500, the image template matching is performed within a 50-meter range in all directions (front, rear, left, and right) centered around the vehicle.For road similarity analysis (Fig.\ref{fig:GetSimiMap}), we utilize a U-Net-based neural network (Fig.\ref{fig:UNet}) to perform image reconstruction. The feature matrix is extracted from the second-to-last layer of the decoder, with dimensions of 500$\times$500$\times$64.

We conducted extensive field experiments in the moderately undulating mountainous terrain of Northwest China, spanning three seasons: summer, autumn, and winter. Furthermore, experiments were conducted under nighttime conditions. Fig.~\ref{fig:SimiContrast} presents satellite map of the same location, alongside BEV images captured across different seasons and at night, along with their corresponding road similarity images. As shown in Fig.~\ref{fig:SimiContrast}, seasonal variations lead to a gradual degradation of road color features in off-road environments, whereas road similarity features exhibit significantly greater robustness. Moreover, under nighttime conditions, the road similarity features remain clearly evident. Thus, by explicitly mapping both the satellite imagery and BEV perception images into the road similarity space, we effectively mitigate the impact of seasonal, regional, and lighting variations on image template matching.

\subsection{Localization Accuracy Assessment and Comparative Analysis}

As depicted in Fig.~\ref{fig:TemplateMatching}, in a straight road segment, the particles are mainly distributed along the vehicle's longitudinal axis. Due to the absence of distinctive longitudinal features, high-weight particles tend to cluster at the image center, forming a vertical red band. This phenomenon indicates reduced localization accuracy along the vehicle's longitudinal axis.This ultimately leads to longitudinal errors in the localization method under such conditions. In contrast, prominent lateral road boundary features effectively constrain the localization results to the road center. To quantify localization accuracy, we introduce two error metrics: Absolute Trajectory Error (ATE) and Lateral Path Error (LPE).

\begin{align}
	ATE &= \frac{1}{N} \sum_{t=1}^{N} \left\| p_{\text{pred}}^{t} - p_{\text{gt}}^{t} \right\| \label{eq:ATE} \\
	LPE &= \frac{1}{N} \sum_{t=1}^{N} \underset{i}{\min} \left( \left| p_{\text{pred}}^{t} - p_{\text{gt}}^{i} \right| \right) \label{eq:LPE}
\end{align}

\noindent where \( N \) is the total number of time steps in the trajectory, \( p_{\text{pred}}^{t} \) represents the predicted position of the system at time step \( t \), \( p_{\text{gt}}^{t} \) denotes the ground truth position at time step \( t \), \( \| p_{\text{pred}}^{t} - p_{\text{gt}}^{t} \| \) denotes the Euclidean distance between the predicted and ground truth positions at time \( t \), \( \underset{i}{\min} \left( \left| p_{\text{pred}}^{t} - p_{\text{gt}}^{i} \right| \right) \) computes the minimum Euclidean distance between the predicted position at time \( t \) and all ground truth positions.

\begin{figure}[htbp]
	\centering
	\includegraphics[width=\columnwidth]{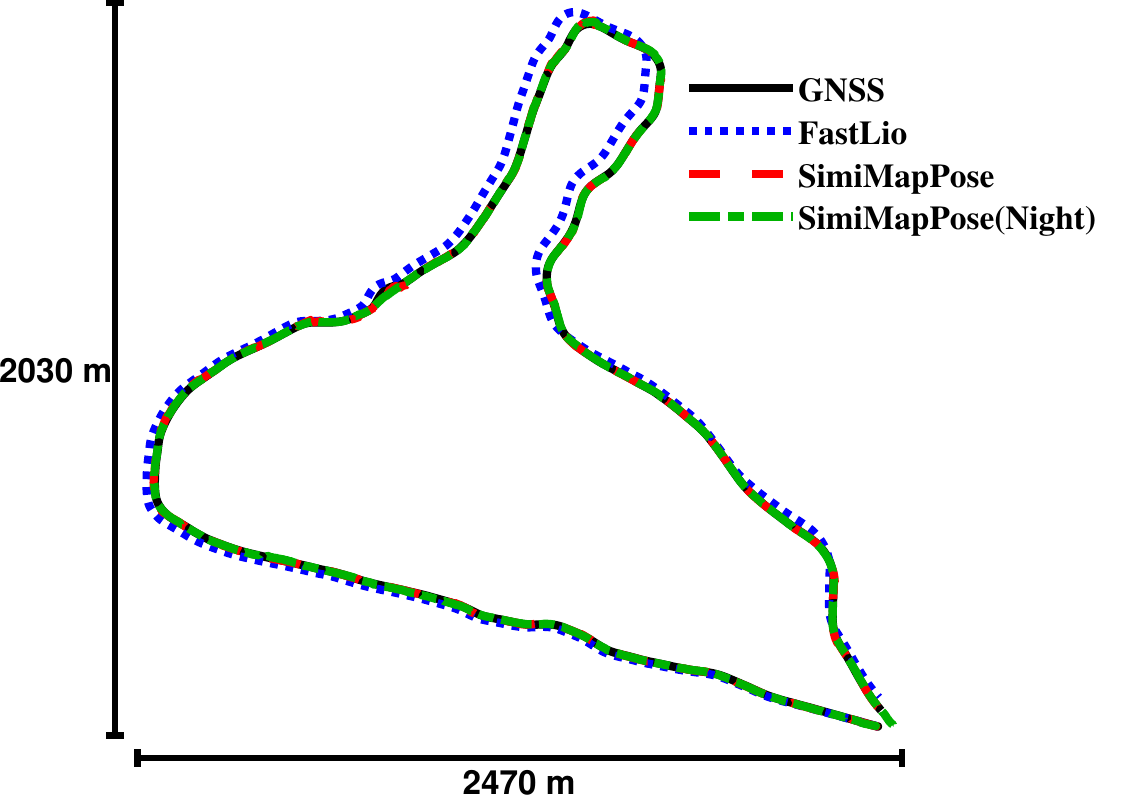}
	\caption{Qualitative Comparison of Localization Methods. The solid black line represents the GNSS positioning result, the blue dashed line corresponds to the odometry-based FastLio results, the red dashed line indicates the localization result of our proposed method under daytime conditions, and the blue dash-dotted line represents the localization result of our method under night-time conditions.}
	\label{fig:trajectory_plot}
\end{figure}

To assess the effectiveness of the proposed method, we compare its localization results against FastLio, which serves as the baseline. Our approach is referred to as SimiMapPose. Experiments were conducted over a 5.2 km trajectory in an off-road environment. Additionally, to further validate the robustness of the proposed method, we performed localization experiments in nighttime conditions and conducted a continuous 10 km long-distance test during the daytime. The qualitative and quantitative comparisons are presented in Fig.~\ref{fig:trajectory_plot} and Table~\ref{tab:Comparison}, respectively.

\begin{table}[H]
	\centering  
	\caption{Quantitative Localization Performance of Different Methods (in Meters)} 
	\label{tab:Comparison} 
	
	\vspace{4pt} 
	
	\setlength{\tabcolsep}{4pt} 
	
	\begin{tabular}{lc>{\bfseries}c>{\bfseries}c>{\bfseries}c}  
		\toprule 
		Metric & FastLio & \textbf{SimiMapPose} & \textbf{\makecell{SimiMapPose\\(night)}} & \textbf{\makecell{SimiMapPose\\(10 km)}} \\ 
		\midrule 
		ATE &  24.47 & \textbf{3.83} & \textbf{2.41} & \textbf{3.41} \\ 
		LPE &  15.63 & \textbf{0.71} & \textbf{0.83} & \textbf{0.89} \\ 
		\bottomrule 
	\end{tabular}
\end{table}

As shown in Fig.~\ref{fig:trajectory_plot}, the proposed method demonstrates strong agreement with the satellite-based ground truth. In contrast, FastLio exhibits progressive localization drift over long trajectories. These results indicate that the proposed method effectively mitigates drift errors inherent in odometry-based localization, thereby improving both accuracy and stability in long-distance navigation. As shown in Table~\ref{tab:Comparison}, the proposed method achieves significantly lower LPE compared to ATE, suggesting that longitudinal deviations are the primary contributor to ATE error. Moreover, SimiMapPose outperforms FastLio in both ATE and LPE metrics, further validating the effectiveness of the proposed approach. Overall, the proposed method maintains high localization accuracy and stability over an extended 10 km trajectory. Notably, the proposed method maintains accurate and stable localization even under nighttime conditions, further demonstrating its robustness and adaptability.

\subsection{Ablation Study}

This section presents an ablation study to validate the effectiveness of the proposed localization method, which operates by projecting both the satellite map and the BEV perception image into the road similarity space. To analyze the contribution of the proposed road similarity space, we introduce an alternative localization method, ColorMapPose, which directly matches the raw BEV perception image with the satellite map in the RGB color space. Experiments were conducted using datasets collected in both summer and winter, covering a total trajectory length of 5.2 km. The quantitative results are presented in Table~\ref{tab:Ablation}.

\begin{table}[H]
	\centering  
	\caption{ABLATION STUDY ON ROAD SIMILARITY MATCHING (IN METERS)} 
	\label{tab:Ablation} 
	\vspace{4pt} 
	\setlength{\tabcolsep}{5pt} 
	\begin{tabular}{lcc>{\bfseries}c}  
		\toprule 
		Metric & \makecell{ColorMapPose\\(Summer)} & \makecell{ColorMapPose\\(Winter)} & \textbf{\makecell{SimiMapPose\\(Winter)}} \\ 
		\midrule 
		ATE & 8.39 & 18.58 & \textbf{3.83} \\  
		LPE & 1.04 & 12.01 & \textbf{0.71} \\  
		\bottomrule 
	\end{tabular} 
\end{table}

As shown in Fig.\ref{fig:SimiContrast} and Table\ref{tab:Ablation}, the performance of direct RGB-based matching between the raw BEV perception image and the satellite map degrades significantly as environmental conditions shift from summer to winter. However, even in winter conditions, localization in the road similarity space consistently surpasses RGB-based localization across all tested seasons. These results validate the effectiveness of the proposed method in constructing a robust road similarity space for localization.

\section{CONCLUSIONS AND FUTURE WORK}

This study introduces a matching-based localization framework that explicitly projects BEV perception images and satellite map into a unified feature space, termed the road similarity space. This approach effectively reduces the sensitivity of feature matching to seasonal, illumination, and other environmental variations. Additionally, by integrating path matching, it enables long-range, robust, and high-precision localization in GNSS-denied environments, outperforming state-of-the-art methods. In contrast, existing approaches typically rely on neural networks to implicitly project satellite map and local vehicle perception results into an embedded feature space, where similarity is assessed using Euclidean distance computation or direct regression of relative position information. The proposed method exhibits superior generalization capability while maintaining enhanced interpretability.

However, the current method is primarily applicable to off-road environments with visible trails. Additionally, in straight-road environments, the absence of distinctive longitudinal features may induce longitudinal drift, thereby increasing real-time localization errors. To enhance localization robustness and accuracy, future work will integrate visual odometry to improve local motion estimation. Furthermore, by quantifying the uncertainty of image matching uncertainty, we will integrate odometry and image matching results into a Bayesian filtering framework to refine global position estimation.

\vfill
\bibliographystyle{IEEEtran}

\bibliography{reference.bib}

\begin{thebibliography}{10}
\providecommand{\url}[1]{#1}
\csname url@samestyle\endcsname
\providecommand{\newblock}{\relax}
\providecommand{\bibinfo}[2]{#2}
\providecommand{\BIBentrySTDinterwordspacing}{\spaceskip=0pt\relax}
\providecommand{\BIBentryALTinterwordstretchfactor}{4}
\providecommand{\BIBentryALTinterwordspacing}{\spaceskip=\fontdimen2\font plus
\BIBentryALTinterwordstretchfactor\fontdimen3\font minus
  \fontdimen4\font\relax}
\providecommand{\BIBforeignlanguage}[2]{{%
\expandafter\ifx\csname l@#1\endcsname\relax
\typeout{** WARNING: IEEEtran.bst: No hyphenation pattern has been}%
\typeout{** loaded for the language `#1'. Using the pattern for}%
\typeout{** the default language instead.}%
\else
\language=\csname l@#1\endcsname
\fi
#2}}
\providecommand{\BIBdecl}{\relax}
\BIBdecl

\bibitem{SLAM}
H.~Durrant-Whyte and T.~Bailey, ``Simultaneous localization and mapping: part
  i,'' \emph{IEEE robotics \& automation magazine}, vol.~13, no.~2, pp.
  99--110, 2006.

\bibitem{DEM}
X.~Wan, Y.~Shao, S.~Zhang, and S.~Li, ``Terrain aided planetary uav
  localization based on geo-referencing,'' \emph{IEEE Transactions on
  Geoscience and Remote Sensing}, vol.~60, pp. 1--18, 2022.

\bibitem{OpenStreetMap}
P.-E. Sarlin, D.~DeTone, T.-Y. Yang, A.~Avetisyan, J.~Straub, T.~Malisiewicz,
  S.~R. Bulo, R.~Newcombe, P.~Kontschieder, and V.~Balntas, ``Orienternet:
  Visual localization in 2d public maps with neural matching,'' in
  \emph{Proceedings of the IEEE/CVF Conference on Computer Vision and Pattern
  Recognition}, 2023, pp. 21\,632--21\,642.

\bibitem{viswanathan2014vision}
A.~Viswanathan, B.~R. Pires, and D.~Huber, ``Vision based robot localization by
  ground to satellite matching in gps-denied situations,'' in \emph{2014
  IEEE/RSJ International Conference on Intelligent Robots and Systems}.\hskip
  1em plus 0.5em minus 0.4em\relax IEEE, 2014, pp. 192--198.

\bibitem{cheung2018cross}
C.~Cheung and S.~Baek, ``Cross correlating ground-level panoramas with
  satellite imagery for gps-denied localization of autonomous ground
  vehicles,'' in \emph{NDIA Ground Vehicle Systems Engineering and Technology
  Symposium}, 2018.

\bibitem{litman2022gps}
Y.~Litman, D.~McGann, E.~Dexheimer, and M.~Kaess, ``Gps-denied global
  visual-inertial ground vehicle state estimation via image registration,'' in
  \emph{2022 International Conference on Robotics and Automation (ICRA)}.\hskip
  1em plus 0.5em minus 0.4em\relax IEEE, 2022, pp. 8178--8184.

\bibitem{niles2022satellite}
K.~Niles, S.~Bunkley, W.~Wagner, I.~Blankenau, A.~Netchaev, and
  A.~Soylemezoglu, ``Satellite image template matching with covariance
  estimation for unmanned ground vehicle localization: Active terrain
  localization imaging system (atlis),'' in \emph{Ground Vehicle Systems
  Engi-neering and Technology Symposium (GVSETS)}, 2022.

\bibitem{hu2018cvm}
S.~Hu, M.~Feng, R.~M. Nguyen, and G.~H. Lee, ``Cvm-net: Cross-view matching
  network for image-based ground-to-aerial geo-localization,'' in
  \emph{Proceedings of the IEEE Conference on Computer Vision and Pattern
  Recognition}, 2018, pp. 7258--7267.

\bibitem{UBEV}
A.~B. Camiletto, A.~Bochicchio, A.~Liniger, D.~Dai, and A.~Gawel, ``U-bev:
  Height-aware bird’s-eye-view segmentation and neural map-based
  relocalization,'' in \emph{2024 IEEE/RSJ International Conference on
  Intelligent Robots and Systems (IROS)}.\hskip 1em plus 0.5em minus
  0.4em\relax IEEE, 2024, pp. 5597--5604.

\bibitem{kim2017satellite}
D.-K. Kim and M.~R. Walter, ``Satellite image-based localization via learned
  embeddings,'' in \emph{2017 IEEE international conference on robotics and
  automation (ICRA)}.\hskip 1em plus 0.5em minus 0.4em\relax IEEE, 2017, pp.
  2073--2080.

\bibitem{fu2020lidar}
M.~Fu, M.~Zhu, Y.~Yang, W.~Song, and M.~Wang, ``Lidar-based vehicle
  localization on the satellite image via a neural network,'' \emph{Robotics
  and Autonomous Systems}, vol. 129, p. 103519, 2020.

\bibitem{2D3D}
L.~Li, Y.~Ma, K.~Tang, X.~Zhao, C.~Chen, J.~Huang, J.~Mei, and Y.~Liu,
  ``Geo-localization with transformer-based 2d-3d match network,'' \emph{IEEE
  Robotics and Automation Letters}, vol.~8, no.~8, pp. 4855--4862, 2023.

\bibitem{boosting3dof}
Y.~Shi, F.~Wu, A.~Perincherry, A.~Vora, and H.~Li, ``Boosting 3-dof
  ground-to-satellite camera localization accuracy via geometry-guided
  cross-view transformer,'' in \emph{Proceedings of the IEEE/CVF International
  Conference on Computer Vision}, 2023, pp. 21\,516--21\,526.

\bibitem{xu2021fast}
W.~Xu and F.~Zhang, ``Fast-lio: {A Fast, Robust LiDAR-Inertial Odometry Package
  by Tightly-Coupled Iterated Kalman Filter},'' \emph{IEEE Robotics and
  Automation Letters}, vol.~6, no.~2, pp. 3317--3324, 2021.

\bibitem{vstrong}
\BIBentryALTinterwordspacing
S.~Jung, J.~Lee, X.~Meng, B.~Boots, and A.~Lambert, ``V-strong: {Visual
  Self-Supervised Traversability Learning for Off-road Navigation},'' 2024.
  [Online]. Available: \url{https://arxiv.org/abs/2312.16016}
\BIBentrySTDinterwordspacing

\bibitem{bigmap}
\BIBentryALTinterwordspacing
BigMap, ``Bigmap: High-resolution satellite map data,'' 2023, accessed:
  2023-10-20. [Online]. Available: \url{http://www.bigemap.com/}
\BIBentrySTDinterwordspacing

\end{thebibliography}
\end{document}